\documentclass[sigconf]{acmart}
\AtBeginDocument{%
  }

\copyrightyear{2026}
\acmYear{2026}
\setcopyright{cc}
\setcctype{by-nc-nd}
\acmConference[WWW '26]{Proceedings of the ACM Web Conference 2026}{April 13--17, 2026}{Dubai, United Arab Emirates}
\acmBooktitle{Proceedings of the ACM Web Conference 2026 (WWW '26), April 13--17, 2026, Dubai, United Arab Emirates}
\acmPrice{}
\acmDOI{10.1145/3774904.3792512}
\acmISBN{979-8-4007-2307-0/2026/04}

\usepackage{multirow}
\usepackage{makecell}
\usepackage[most]{tcolorbox}
\usepackage{tabularx}
\usepackage{xcolor}
\usepackage{float}
\usepackage{subcaption}
\usepackage{balance}

\usepackage{enumitem}

\newtcolorbox{examplebox}[1][]{
  colback=gray!5!white,
  colframe=black!20,
  fonttitle=\bfseries,
  title=#1,
  boxrule=0.3pt,
  arc=2pt,
  left=8pt,
  right=8pt,
  top=6pt,
  bottom=6pt,
  enhanced,
}
\begin{document}

\title{\textsc{CompactRAG}: Reducing LLM Calls and Token Overhead in Multi-Hop Question Answering}


\author{Hao Yang}
\authornote{All authors contributed equally to this research.}
\affiliation{%
  \institution{State Key Laboratory for Novel Software Technology, Nanjing University}
  \city{Suzhou}
  \state{Jiangsu}
  \country{China}
}
\email{howyoung80@163.com}

\author{Zhiyu Yang}
\authornotemark[1]
\affiliation{%
  \institution{Erik Jonsson School of Engineering and Computer Science, University of Texas at Dallas}
  \city{Richardson}
  \state{Texas}
  \country{USA}
}
\email{zhiyu.yang@utdallas.edu}

\author{Xupeng Zhang}
\authornotemark[1]
\affiliation{%
  \institution{Isoftstone Information Technology (Group) Co.,Ltd.}
  \city{Beijing}
  \country{China}
}
\email{lagelangpeng@gmail.com}

\author{Wei Wei}
\affiliation{%
  \institution{College of Electronic and Information Engineering, Tongji University }
  \city{Shanghai}
  \country{China}
}
\email{2510856@tongji.edu.cn}

\author{Yunjie Zhang}
\affiliation{
  \institution{School of Electronic Information, Central South University}
  \city{Changsha}
  \state{Hunan}
  \country{China}
}
\email{Zhangyj@csu.edu.cn}

\author{Lin Yang}
\authornote{Corresponding author.}
\orcid{0000-0001-9056-0500}
\affiliation{
  \institution{State Key Laboratory for Novel Software Technology, Nanjing University}
  \city{Suzhou}
  \state{Jiangsu}
  \country{China}
}
\email{linyang@nju.edu.cn}

\renewcommand{\shortauthors}{Hao Yang et al.}







\renewcommand{\shortauthors}{Yang et al.}

\begin{abstract}
Retrieval-augmented generation (RAG) has become a key paradigm for knowledge-intensive question answering.
However, existing multi-hop RAG systems remain inefficient, as they alternate between retrieval and reasoning at each step, resulting in repeated LLM calls, high token consumption, and unstable entity grounding across hops.
We propose \textsc{CompactRAG}, a simple yet effective framework that decouples offline corpus restructuring from online reasoning.

In the offline stage, an LLM reads the corpus once and converts it into an \emph{atomic QA knowledge base}, which represents knowledge as minimal, fine-grained question--answer pairs.
In the online stage, complex queries are decomposed and carefully rewritten to preserve entity consistency, and are resolved through dense retrieval followed by RoBERTa-based answer extraction.
Notably, during inference, the LLM is invoked only twice in total—once for sub-question decomposition and once for final answer synthesis—regardless of the number of reasoning hops.

Experiments on \textsc{HotpotQA}, \textsc{2WikiMultiHopQA}, and \textsc{MuSiQue} demonstrate that \textsc{CompactRAG} achieves competitive accuracy while substantially reducing token consumption compared to iterative RAG baselines, highlighting a cost-efficient and practical approach to multi-hop reasoning over large knowledge corpora.
The implementation is available at \url{https://github.com/How-Young-X/CompactRAG}.

\end{abstract}

\ccsdesc[500]{Information systems~Retrieval models and ranking}
\ccsdesc[300]{Computing methodologies~Natural language processing}

\keywords{Retrieval-augmented generation, Multi-hop question answering, Efficient reasoning}



\maketitle


\section{Introduction}

Retrieval-Augmented Generation (RAG)~\cite{rag} is now a standard approach for knowledge intensive NLP.  
RAG combines explicit retrieval with the generation and reasoning capacity of large language models (LLMs).  
This combination works well for question answering. LLMs can produce factual, grounded answers by retrieving relevant passages~\cite{karpukhin-etal-2020-dense,gao2024rag,luo2023dricldemonstrationretrievedincontextlearning, atlas,borgeaud2022improving}. However, multi-hop question answering (MHQA)~\cite{yang-etal-2018-hotpotqa,trivedi2021musique,xanh2020_2wikimultihop,tang2024multihoprag,meqa} is more challenging.  
A MHQA query requires integrating evidence from multiple documents.  
Conventional RAG pipelines face three recurring problems in this setting.  
First, efficiency degrades as reasoning hops increase~\cite{fang-etal-2024-trace,ji2025curriculumguidedreinforcementlearning,tang2024multihoprag}.  
Second, retrieved context often contains redundant information~\cite{shi2024retrieval,jin2025sara,saleh2025sg,rawte2025radiantretrievalaugmentedentitycontext}.  
Third, maintaining factual consistency across hops is difficult~\cite{jimenez2024hipporag,zhong-etal-2023-mquake,guo-etal-2023-counterfactual}. These challenges are central to web scale information access, where large and heterogeneous knowledge sources must be efficiently retrieved, represented, and reasoned over by LLMs.

Recent work implements iterative retrieval generation cycles for multi-hop reasoning.  
Examples include \textsc{Self-Ask}~\cite{selfask}, \textsc{IRCoT}~\cite{ircot}, and \textsc{Iter-RetGen}~\cite{itergen}.  
These methods alternate between retrieval and LLM reasoning.  
At each step, the model retrieves passages guided by prior reasoning.  
This design improves factual coverage and yields explicit reasoning chains.  
It also increases the number of LLM invocations.  
As a result, token usage and latency grow with hop depth.  
This growth raises computational cost and limits scalability. Multi-hop question decomposition can also harm retrieval accuracy, a common failure mode is \emph{entity drift}. A decomposed sub-question may lose its explicit entity mention.  
For example, ``Where was the scientist who discovered penicillin born?'' can be split into ``Who discovered penicillin?'' and ``Where was \emph{he} born?'' The second sub-question lacks an explicit entity and may retrieve unrelated documents, producing inconsistent results~\cite{ChainRAG}.  
Prior work documents related failures when decomposition is imprecise~\cite{ChainRAG,perez2020unsupervisedquestiondecompositionquestion}.

A large body of follow up work attempts to mitigate these issues by refining the retrieval–reasoning interaction. \textsc{HopRAG}~\cite{hoprag} and \textsc{LevelRAG}~\cite{levelrag} introduce hierarchical or logic-aware retrieval to enhance reasoning paths, yet still require multiple LLM invocations per hop. \textsc{DualRAG}~\cite{dualrag} and \textsc{GenGround}~\cite{grounded} employ iterative “generate then ground’’ loops to couple generation and retrieval, which increases computational overhead. \textsc{Q-DREAM}~\cite{qdream} dynamically optimizes sub-question semantics in a learned retrieval space, but depends on several LLM-driven refinement stages. \textsc{ChainRAG}~\cite{ChainRAG} builds a sentence-level graph to preserve entity continuity and alleviate entity drift, at the cost of heavy graph traversal and multiple reasoning–retrieval cycles. Other works leverage internal model signals such as attention entropy or decoding uncertainty~\cite{i0,i1,i2,i3,i4}, but these approaches require access to non-public model activation, limiting their deployability. Finally, \textsc{EfficientRAG}~\cite{efficientrag} reduces online LLM involvement via lightweight retriever modules, yet still operates directly over raw corpus passages, leaving substantial redundancy in retrieved context.

We propose \textsc{CompactRAG}, a simple and practical alternative.  
CompactRAG separates corpus processing from online inference.  
Offline, an LLM reads the corpus once and constructs an \emph{atomic QA knowledge base}.  
These QA pairs are concise, fact-level units that reduce redundancy and better align with question semantics~\cite{tan2024blinded}.  
Online, a complex query is decomposed into dependency-ordered sub-questions.  
Each sub-question is resolved using lightweight modules for retrieval, answer extraction, and question rewriting.  
The main LLM is invoked only twice per query: once for decomposition and once for final synthesis.  
This fixed two-call design makes LLM usage independent of hop depth.  
The offline step incurs a one-time cost.  
That cost is amortized as user queries accumulate.  



\textbf{Contributions.} Our work makes three main contributions. First, we analyze scalability issues in iterative RAG pipelines, showing how token consumption and LLM calls grow with reasoning depth. Second, we introduce \textsc{CompactRAG}, a two-call RAG framework that uses an offline atomic QA knowledge base and lightweight online modules to enable efficient multi-hop inference. Third, we evaluate CompactRAG on \textsc{HotpotQA}, \textsc{2WikiMultiHopQA}, and \textsc{MuSiQue}, demonstrating competitive accuracy along with significant reductions in inference token usage compared to strong iterative baselines.

\section{Related Work}

We review related work in three main areas: (1) multi-hop question answering and iterative retrieval–reasoning pipelines, (2) structured and corpus-level retrieval enhancement, and (3) efficiency oriented and adaptive retrieval strategies.  
Our discussion highlights how \textsc{CompactRAG} differs from these paradigms by decoupling reasoning from retrieval through an offline–online architecture.
\subsection{Multi-hop QA and Iterative Retrieval–Reasoning Pipelines}
Multi-hop question answering benchmarks such as \textsc{HotpotQA}~\cite{yang-etal-2018-hotpotqa}, \textsc{2WikiMultiHopQA}~\cite{xanh2020_2wikimultihop}, and \textsc{MuSiQue}~\cite{trivedi2021musique} have driven research on compositional reasoning across documents.  
Early retrieval augmented approaches treat reasoning as a sequence of retrieval and generation steps.  
\textsc{Self-Ask}~\cite{selfask} explicitly decomposes questions into sub-questions that are answered iteratively, using the model’s own intermediate outputs as guidance.  
\textsc{IRCoT}~\cite{ircot} interleaves retrieval with a chain-of-thought process~\cite{cot}, allowing reasoning traces to refine retrieval queries dynamically.  
\textsc{Iter-RetGen}~\cite{itergen} further integrates iterative retrieval and generation, where each model response serves as a context for the next retrieval round.  
While these systems enhance factual completeness and interpretability, their reliance on repeated LLM invocations makes computational cost scale linearly with reasoning hops.  
Each iteration expands the prompt with retrieved passages, leading to excessive token consumption and increased latency.  
Moreover, automatic sub-question decomposition can suffer from \emph{entity drift}~\cite{perez2020unsupervisedquestiondecompositionquestion,ChainRAG}, where referential grounding is lost (e.g., ``Where was \emph{he} born?''), degrading retrieval precision.  
\textsc{CompactRAG} eliminates these iterative dependencies by executing retrieval and reasoning separately, using fixed-cost local modules for sub-question resolution.

\subsection{Structured and Corpus-level Retrieval Enhancement}
Beyond iterative pipelines, several studies improve retrieval grounding by introducing explicit structure or corpus level representations.  
\textsc{HopRAG}~\cite{hoprag} constructs paragraph graphs linking documents through logical dependencies, enabling LLM-guided traversal across hops.  
\textsc{LevelRAG}~\cite{levelrag} employs a hierarchical planner that combines sparse, dense, and web based retrieval to support multi-hop reasoning.  
\textsc{DualRAG}~\cite{dualrag} and \textsc{GenGround}~\cite{grounded} couple generation and retrieval through dual or generate then ground loops, progressively refining sub-queries.  
However, these designs require multiple LLM calls for reasoning validation and query reformulation, limiting efficiency.  
\textsc{Q-DREAM}~\cite{qdream} learns a dynamic retrieval space aligned to sub-question semantics using LoRA-tuned modules, while \textsc{ChainRAG}~\cite{ChainRAG} constructs sentence-level graphs to maintain entity continuity and mitigate lost-in-retrieval errors.  
Although such structures improve reasoning fidelity, they often entail costly graph traversal, embedding computation, and repeated model inference.  

Another direction focuses on corpus preprocessing.  
\textsc{EfficientRAG}~\cite{efficientrag} introduces lightweight modules—\textit{Labeler}, \textit{Tagger}, and \textit{Filter}—to reduce online LLM calls, but it still retrieves over raw, redundant passages.  
Recent studies~\cite{tan2024blinded} observe that LLM-generated text aligns more closely with the query’s semantic space and thus serves as a more compact and expressive retrieval unit.  
Inspired by this, \textsc{CompactRAG} performs offline corpus restructuring into atomic QA pairs.  
This produces semantically complete, redundancy-free, and fact-centric knowledge units that support fine-grained reasoning.  
Unlike prior structural frameworks, CompactRAG requires no online graph traversal or dynamic refinement, maintaining retrieval efficiency and stable accuracy.
\subsection{Efficiency and Adaptive Retrieval Strategies}
A complementary line of research improves efficiency through adaptive retrieval or model-aware decision mechanisms.  
\textsc{DioR}~\cite{i3}, \textsc{SeaKR}~\cite{i2}, and \textsc{DRAGIN}~\cite{i4} propose adaptive retrieval-augmented generation  methods that monitor model internal signals, such as entropy, gradient variance, or decoding uncertainty to determine when to retrieve additional context.  
Active RAG~\cite{i0} and Entropy-Based Decoding ~\cite{i1} follow similar strategies, activating retrieval only when confidence drops.  
Although effective in reducing redundant retrievals, these systems require access to hidden activations or attention scores, which are typically unavailable in closed weight LLMs, restricting their practicality.  
In contrast, \textsc{CompactRAG} achieves comparable efficiency gains through architectural design rather than internal signal access.  
Its offline–online separation amortizes reasoning cost across queries: the knowledge base is built once offline, and each query requires only lightweight retrieval and two fixed LLM calls online.  
This design ensures predictable cost, scalability, and compatibility with open or closed LLMs.

\paragraph{Summary.}
Existing RAG systems trade off reasoning accuracy, retrieval precision, and computational efficiency.  
Iterative pipelines improve factual reasoning but scale poorly with hop depth; graph-based and dynamic retrieval methods enhance grounding but require complex online computation; internal signal approaches remain difficult to deploy.  
\textsc{CompactRAG} reconciles these limitations by precomputing atomic QA representations offline and performing reasoning through modular, low cost components online.  
This results in a scalable and token efficient framework for multi-hop reasoning, achieving a favorable balance between accuracy and efficiency.

\section{Methodology}
\label{sec:methodology}

The goal of \textsc{CompactRAG} is to reduce corpus redundancy, minimize token consumption during complex reasoning, and decrease the number of LLM calls required for multi-hop question answering. To this end, we decompose the reasoning process into two stages: (1) an \emph{offline corpus preprocessing stage}, which constructs a concise and structured  QA knowledge base, and (2) an \emph{online reasoning stage}, which efficiently retrieves and composes relevant atomic QA pairs without repeated LLM invocations.

In the offline stage, the raw corpus is processed once by an LLM to generate a compact set of atomic QA pairs, removing noise and redundancy. In the online stage, a complex question is decomposed into a dependency graph of sub-questions, which are then iteratively resolved through retrieval over the QA knowledge base. The retrieved evidence is aggregated for a single final LLM call that synthesizes the final answer. This design keeps the number of LLM calls fixed and ensures efficiency even for deep multi-hop reasoning chains.

\subsection{Offline Stage}

\begin{figure}[htbp]
    \centering
    \includegraphics[width=0.8\linewidth, height=8cm]{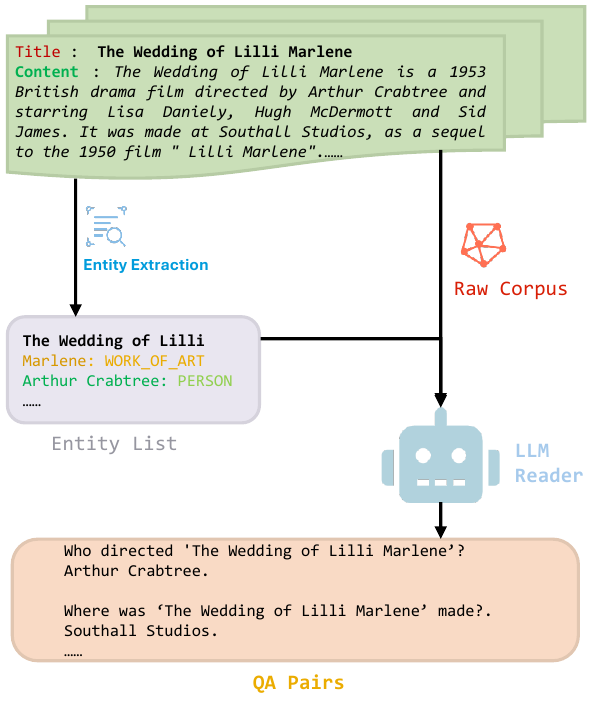}
    \caption{
    \textbf{Overview of the Offline Knowledge Construction in \textsc{CompactRAG}.}
    The raw corpus is first processed by an LLM ``Reader'' that reformulates document content into a set of atomic QA pairs.  
    Each QA pair captures a minimal factual unit, annotated with entity information to ensure semantic completeness and prevent redundancy.  
    }
    \label{fig:reader}
\end{figure}

In the offline preprocessing stage, \textsc{CompactRAG} employs an LLM to transform the raw corpus into a structured and compact \emph{atomic QA knowledge base}.  
This process is performed once prior to inference, aiming to eliminate redundancy while preserving essential factual information in a form directly aligned with downstream query semantics.  
Inspired by prior findings~\cite{tan2024blinded} that LLM-generated representations tend to align more closely with the semantic space of natural queries, we prompt the LLM to read each document and reformulate its content into a set of \emph{atomic QA pairs}.  
Each pair expresses a single factual statement with minimal granularity, ensuring non-overlapping information units suitable for multi-hop composition.  
Before generation, entities within the corpus are automatically annotated using \texttt{SpaCy}\footnote{\url{https://spacy.io}}, and these entities are explicitly enforced in the generation prompt (see Appendix~\ref{prompt:qa}).  
This constraint guarantees semantic completeness and prevents omission of key referential elements.  
An overview of this corpus to QA transformation is illustrated in Figure~\ref{fig:reader}.

\vspace{4pt}
\noindent\textbf{Dense retrieval over atomic QA knowledge.}  
After generation, each atomic QA pair is embedded into a shared semantic space using dense retrieval representations.  
Unlike sparse lexical retrieval methods such as BM25, dense retrieval captures contextual and semantic similarity beyond surface word overlap, which is particularly critical for multi-hop reasoning where sub-questions often differ lexically from supporting knowledge.  
To maximize the semantic coherence between questions and answers, the question (\(q\)) and answer (\(a\)) components of each pair are concatenated into a single text segment \([q; a]\) before encoding.  
This joint representation preserves the full factual scope of each unit, allowing the retriever to index both the intent expressed in the question and the corresponding factual content in the answer.  
During online inference, the same encoder retrieves top-\(k\) relevant QA pairs for each sub-question based on embedding similarity, enabling compact and semantically aligned evidence retrieval from the preprocessed knowledge base.

 \begin{figure*}[htbp]
    \centering
    \includegraphics[width=0.98\linewidth, height=9.9cm]{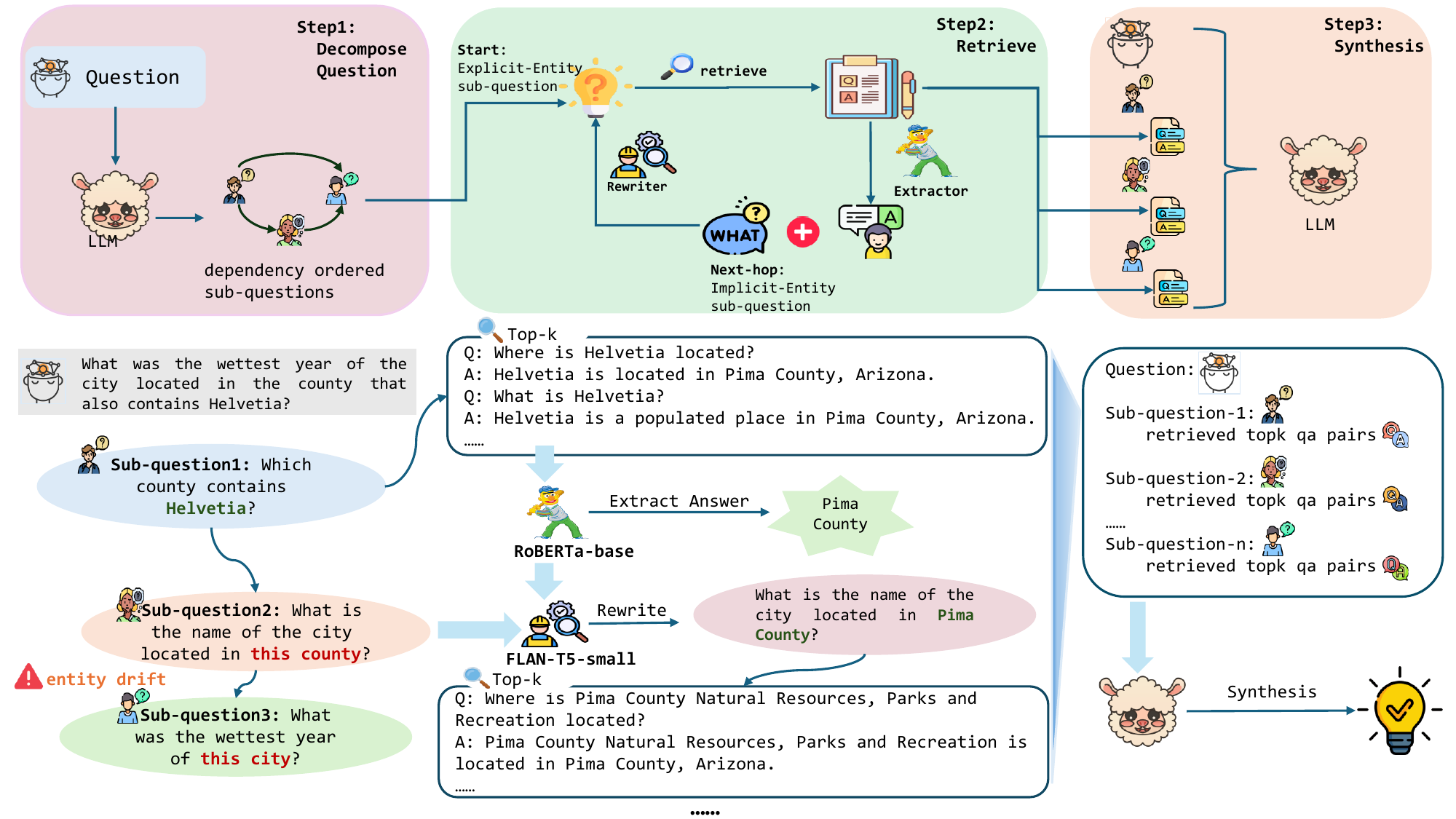}
    \caption{
    \textbf{Overview of the Online Reasoning Pipeline in \textsc{CompactRAG}.}
    The framework begins with query decomposition, where a complex multi-hop question is decomposed into dependency ordered sub-questions.  
    Each sub-question is resolved through iterative retrieval over the atomic QA knowledge base, followed by lightweight answer extraction and question rewriting modules that ensure entity continuity and semantic grounding.  
    Once all sub-questions are resolved, the retrieved QA pairs are aggregated and passed to a final synthesis reasoning step, completing the inference process with only two LLM calls per query.
    }
    \label{fig:framework}
\end{figure*}
\subsection{Online Stage}
As illustrated in Figure~\ref{fig:framework}, the online reasoning process begins by decomposing a complex multi-hop question into a dependency-ordered set of sub-questions. One sub-question’s resolution may depend on the answer to a preceding one. Existing iterative RAG systems perform retrieval and reasoning alternately, invoking the LLM at every step to maintain accuracy, but at the cost of excessive computation and token usage.  
In contrast, \textsc{CompactRAG} leverages the atomic QA knowledge base to decouple retrieval from reasoning entirely. Two lightweight transformer-based modules are introduced—an \emph{Answer Extractor} and a \emph{Sub-Question Rewriter}. These modules enable multi-hop retrieval without involving the LLM, thereby reducing computational overhead and preventing \emph{entity drift} across hops.
\subsubsection{Multihop Question Decomposition}
Given a user query \(Q\), the system first decomposes it into a sequence of sub-questions \(\{q_1, q_2, ..., q_n\}\) organized in a dependency graph \(\mathcal{G}\).  
Each directed edge \(q_i \rightarrow q_j\) indicates that the answer to \(q_i\) is required to resolve \(q_j\).  
The decomposition is performed by an LLM once during inference initialization, and the dependency graph guides the iterative retrieval pipeline.

\subsubsection{Answer Extractor}
The Answer Extractor is responsible for identifying the correct entity or fact from retrieved QA pairs that correspond to a given sub-question \(q_i\).  
Given \(q_i\) and its retrieved candidate QA pairs \(\mathcal{P}_i = \{(q_{i,k}, a_{i,k})\}\), the extractor predicts the start and end positions of the correct answer span within the text context. The learning objective is a span prediction loss defined as:

\begin{equation}
\mathcal{L}_{\mathrm{extract}} = -\frac{1}{N}\sum_{i=1}^{N}\big(\log P(s_i\mid q_i,C_i) + \log P(e_i\mid q_i,C_i)\big),
\end{equation}

where \(C_i\) denotes the concatenation of candidate QA pairs, and \(s_i, e_i\) are the gold start and end token indices.

\paragraph{Training Data.}
To construct supervision for the extractor, we sample source passages from the training splits of the benchmarks used in this paper.  
For each passage, an LLM is prompted to generate sub-questions and corresponding correct and distractor QA pairs.  
The correct answer span is explicitly marked within the gold QA pair, while distractors introduce realistic retrieval noise.

\begin{tcolorbox}[title=Training Example of Answer Extractor]
\textbf{Sub-question:} ``Which country is the Eiffel Tower located in?''\\[6pt]
\textbf{QA pairs:}\\[2pt]
\begin{tabularx}{\linewidth}{@{}p{1em}Xr@{}}
(1) & ``Where is the Eiffel Tower situated?''  \textit{``Paris, France''} & \textbf{\textcolor{green!50!black}{[gold]}}\\
(2) & ``What is the height of the Eiffel Tower?''  \textit{``324 meters''} & \textcolor{gray!70!black}{[distractor]}\\
(3) & ``Which city hosts the Colosseum?''  \textit{``Rome, Italy''} & \textcolor{gray!70!black}{[distractor]}\\
\end{tabularx}\\[4pt]
\textbf{Target span:} ``France''
\end{tcolorbox}

The model learns to select the precise supporting evidence under noisy retrieval conditions.
\subsubsection{Sub-Question Rewriter}
As sub-questions may contain ambiguous or coreferential expressions (e.g., pronouns such as ``he'' or ``it''), the Sub-Question Rewriter reformulates the current sub-question \(q_{i+1}\) by explicitly grounding it with the answer entity extracted from the preceding sub-question \(a_i\).  
This mechanism ensures entity continuity across reasoning hops and prevents semantic drift during multi-hop retrieval.

\paragraph{Training Data.}
The data construction process mirrors the extractor setup. For each sample, the LLM generates an ambiguous question, an entity that resolves the ambiguity, and a corresponding rewritten form. To enhance robustness, additional samples are created through controlled perturbations such as entity masking and pronoun insertion.

\begin{tcolorbox}[title=Training Example of Question Rewrite]
\textbf{Ambiguous question:} ``Where was he born?''\\
\textbf{Entity (from previous answer):} ``Albert Einstein''\\
\textbf{Rewritten question:} ``Where was Albert Einstein born?''
\end{tcolorbox}

The rewriter is trained using a conditional generation objective:
\begin{equation}
\mathcal{L}_{\mathrm{rewrite}} = -\frac{1}{N}\sum_{i=1}^{N}\sum_{t=1}^{T_i}\log P(w_{i,t}\mid w_{i,<t}, q_{{\text{amb}},i}, e_i),
\end{equation}
where \(e_i\) denotes the grounding entity and \(w_{i,t}\) are the tokens of the target rewritten question.  
Teacher forcing is employed to stabilize sequence generation.
\subsubsection{Synthesis Reasoning}
After all sub-questions have been resolved and their supporting QA pairs collected, the system aggregates the retrieved knowledge and dependency chain.  
The final LLM call takes as input:
\[
\{Q, \{q_i, a_i, \mathcal{P}_i\}_{i=1}^n\}
\]
and generates the final answer through holistic reasoning.  
This single synthesis step completes the inference process, ensuring that the total number of LLM calls remains constant—\emph{two per query}, regardless of hop count.
\subsection{Inference Integration}
The full online reasoning workflow proceeds as follows:

\begin{enumerate}
  \item Decompose the complex query \(Q\) into dependency-ordered sub-questions \(\{q_1, q_2, ..., q_n\}\).
  \item For each \(q_i\), retrieve candidate QA pairs \(\mathcal{P}_i\) from the atomic QA knowledge base.
  \item Run the \emph{Answer Extractor} on \((q_i, \mathcal{P}_i)\) to obtain the grounded entity or answer \(a_i\).
  \item Use \(a_i\) to rewrite the next sub-question \(q_{i+1}\) via the \emph{Sub-Question Rewriter}, obtaining \(q_{i+1}^{\text{rew}}\).
  \item Continue until all sub-questions are resolved; aggregate all evidence for final LLM synthesis reasoning.
\end{enumerate}

This modular pipeline effectively decouples retrieval from LLM reasoning, maintaining accuracy and grounding while achieving significant reductions in token consumption and LLM invocations.

\section{Experiment Setup}

\subsection{Benchmarks}
We evaluate CompactRAG on three widely used multi-hop question answering benchmarks: \textsc{HotpotQA}~\cite{yang-etal-2018-hotpotqa}, \textsc{2WikiMultiHopQA}~\cite{xanh2020_2wikimultihop}, and the answerable subset of \textsc{MuSiQue}~\cite{trivedi2021musique}. 
For \textsc{HotpotQA}, we adopt the \emph{distractor setting}, where each question is paired with ten Wikipedia paragraphs, two containing gold supporting facts and eight serving as distractors. 
For \textsc{2WikiMultiHopQA} and \textsc{MuSiQue}, which were originally designed for reading comprehension or mixed settings, we repurpose their associated contexts as the retrieval corpus to fit our evaluation framework. 
Illustrative examples from each benchmark are shown in Table~\ref{tab:benchmark_examples}.

Due to computational constraints and the substantial cost of LLM inference, we uniformly sample 250 questions from the development set of each dataset for evaluation. 
Sampling is performed while preserving the original distribution of question types and reasoning difficulty levels to ensure statistical representativeness. 
The sampled questions constitute our test set, and all corresponding contexts are included in the retrieval corpus used during inference.

\begin{table*}[t]
\centering
\small
\caption{Examples from the three multi-hop QA benchmarks used in our experiments. Each question requires reasoning over multiple Wikipedia paragraphs to arrive at the final answer.}
\label{tab:benchmark_examples}
\begin{tabular}{p{0.15\linewidth}p{0.75\linewidth}}
\toprule
\textbf{Benchmark} & \textbf{Example Question and Reasoning Description} \\
\midrule
\textsc{HotpotQA}&
\textbf{Example:} “Were both the film \emph{Twelve Monkeys} and the TV series it inspired produced by the same company?”  
\newline
\textbf{Reasoning:} \textit{The model must first find that \emph{Twelve Monkeys} (film) was produced by Universal Pictures, then check the producer of the TV adaptation, verifying both were indeed produced by the same studio}. \\
\midrule
\textsc{2WikiMultiHopQA} &
\textbf{Example:} “Who was born earlier, the author of \emph{Pride and Prejudice} or the composer of \emph{The Magic Flute}?”  
\newline
\textbf{Reasoning:} \textit{The model needs to identify that Jane Austen wrote \emph{Pride and Prejudice} and Wolfgang Amadeus Mozart composed \emph{The Magic Flute}, then compare their birth years.} \\
\midrule
\textsc{MuSiQue}&
\textbf{Example:} “Which actor who played a character named Jack also starred in the film \emph{The Departed}?”  
\newline
\textbf{Reasoning:} \textit{Requires multi-step reasoning: find that Jack Dawson was played by Leonardo DiCaprio in \emph{Titanic}, then confirm DiCaprio also starred in \emph{The Departed}}. \\
\bottomrule
\end{tabular}
\end{table*}

\subsection{Evaluation Metrics}
We evaluate our approach from both \textit{accuracy} and \textit{efficiency} perspectives.  
For answer correctness, three complementary metrics are employed: \textbf{Exact Match (EM)}, \textbf{F1}, and \textbf{LLM-based Accuracy (Acc)}.  
\textbf{EM} measures the percentage of predictions that exactly match the gold answer string.  
\textbf{F1} captures the token level overlap between the prediction and the reference, balancing precision and recall.  
However, lexical metrics may underestimate semantically correct responses.  
To address this, we further adopt \textbf{LLM-based Accuracy (Acc)}, in which a strong evaluator LLM assesses whether the predicted answer is semantically consistent with the reference answer, prompt shown as in Appendix~\ref{prompteval}  :


Beyond correctness, we also report the average token consumption per query, counting both input and output tokens during inference.  
This efficiency metric directly reflects computational and monetary cost under real-world deployment, and demonstrates the advantage of our method in reducing redundancy and improving inference efficiency.

\subsection{Baselines}
To evaluate \textsc{CompactRAG}, we compare it against representative retrieval–generation frameworks for multi-hop reasoning, as well as a vanilla RAG baseline.

\paragraph{Vanilla RAG.}
A standard retrieval-augmented generation pipeline that retrieves the top-$k$ passages using the original multi-hop question, followed by a single LLM call for answer generation. This simple one-shot approach lacks explicit reasoning decomposition and serves to highlight the benefits of iterative reasoning and structured query decomposition.

\paragraph{Self-Ask}~\cite{selfask}.
A prompting-based method that enhances chain-of-thought reasoning by allowing the model to ask and answer intermediate questions before producing the final answer. In our implementation, the original search engine is replaced with our retriever for consistent retrieval conditions.

\paragraph{IRCoT}~\cite{ircot}.
An interleaved reasoning–retrieval method that alternates between chain-of-thought generation and retrieval. Each reasoning step guides retrieval toward relevant evidence, while retrieved content refines subsequent reasoning, enabling progressive evidence accumulation.

\paragraph{Iter-RetGen}~\cite{itergen}.
A recent iterative retrieval–generation framework. At each step, the model generates a partial response from the current context, identifies information gaps or unresolved entities, and converts them into new retrieval queries. Retrieved passages are appended, and the model updates its response. We use 4 iterations, following the original paper’s observation that performance saturates after 4 steps.

All methods use the same retrieval corpus and retriever. At each step, the top-5 passages by similarity score are selected as evidence for subsequent reasoning.

\subsection{Models}
We use \textsc{LLaMA3.1-8B}~\cite{llama3} as the main LLM for all baselines and \textsc{CompactRAG}, with decoding temperature set to 0 for deterministic inference.

The \textit{Answer Extractor} is based on \textsc{RoBERTa-base}~\cite{roberta} (125M parameters) and identifies answer spans from retrieved QA pairs. The \textit{Sub-Question Rewriter} uses \textsc{Flan-T5-small}~\cite{flan-t5} (80M parameters) to rewrite ambiguous sub-questions by explicitly inserting resolved entities. Both modules are lightweight, enabling local reasoning without invoking the main LLM.

Training data for these modules are generated with \textsc{GPT-4}~\cite{openai2024gpt4technicalreport} at temperature 0 to ensure precise supervision. For dense retrieval, we adopt \textsc{Contriever}~\cite{izacard2021contriever}, an unsupervised contrastive dense retriever that encodes questions and passages into a shared semantic space, providing robust zero-shot retrieval across domains.

For an upper-bound comparison, we include a variant of \textsc{CompactRAG} where the offline atomic QA knowledge base is constructed directly from the corpus using \textsc{GPT-4}, serving as a reference for evaluating the quality of generated atomic QA knowledge.

\begin{table*}[htbp]
\centering
\caption{
Main results on multi-hop QA benchmarks. 
All methods share identical retrieval settings for fair comparison. 
``Token / Sample'' denotes the average total tokens consumed (prompt + generation) per query during inference. 
Best results are \textbf{bolded}.
}
\label{tab:main_results}
\renewcommand{\arraystretch}{1.05}
\setlength{\tabcolsep}{7pt}
\begin{tabular}{lcccccccccc}
\toprule
\multirow{2}{*}{\textbf{Method}} & 
\multicolumn{3}{c}{\textbf{HotpotQA}} & 
\multicolumn{3}{c}{\textbf{2WikiMultiHopQA}} & 
\multicolumn{3}{c}{\textbf{MuSiQue}} &
\multirow{2}{*}{\textbf{Token / Sample}} \\
\cmidrule(lr){2-4} \cmidrule(lr){5-7} \cmidrule(lr){8-10}
 & EM & F1 & Acc & EM & F1 & Acc & EM & F1 & Acc &  \\
\midrule

Vanilla-RAG & 27.60 & 30.32 & 50.80 & 20.80 & 24.38 & 72 & 5.600 & 11.36 & 8.400 & 2.7K \\
\midrule
Self-Ask         & 23.60 & 26.30 & 40.80 & 27.60 & 33.08 &  34.40 & 19.60 & 28.34 & 24.80 & 6.9K \\
IRCoT            & 42.80 & 48.95 & 65.20 & 42.80 & 48.99 & 48.80 & 21.20 &  29.08 & 32.40 & 10.2K \\
Iter-RetGen      &  \underline{46.80} & 52.24 &  \underline{72.40} & \textbf{50.80} & \textbf{59.73} & \textbf{61.20} & 24.80 & 32.42 & 40.00 & 4.7K \\
\midrule
\textbf{\makecell[l]{CompactRAG\\LLaMA-3.1-8b-Reading\\ (Ours)}} & 45.20 & \underline{66.21} & 70.40 & 40.40 & 49.62 & 53.20 & \underline{26.80} & \underline{37.63} & \underline{41.20} & \textbf{1.9K} \\

\\
\textbf{\makecell[l]{CompactRAG\\GPT-4-Reading\\ (Ours)}} & \textbf{49.60} & \textbf{69.54} & \textbf{77.20} & \underline{47.20} & \underline{55.67} & \underline{57.20} & \textbf{30.80} & \textbf{42.34} & \textbf{43.60} & \textbf{1.9K} \\
\bottomrule
\end{tabular}
\end{table*}

\section{Results and Analysis}
\label{sec:results}

In this section, we present and analyze the experimental results of \textsc{CompactRAG} and competing baselines across three multi-hop QA benchmarks: \textsc{HotpotQA}, \textsc{2WikiMultiHopQA}, and \textsc{MuSiQue}.  
We report Exact Match (EM), F1, and accuracy (Acc) scores, along with the average token consumption per query.  
Results are summarized in Table~\ref{tab:main_results}.

\subsection{Overall Performance}
Table~\ref{tab:main_results} presents the main experimental results, comparing two configurations of \textsc{CompactRAG} against several baseline methods. To ensure a controlled evaluation, all methods utilize the same retrieval setup. Our primary comparison employs the same backbone LLM \textsc{LLaMA-3.1-8B} across methods. Under this setting, \textsc{CompactRAG} achieves competitive performance in accuracy on the multi-hop benchmarks HotpotQA, 2WikiMultiHopQA, and MuSiQue. Notably, it attains this performance while consuming significantly fewer tokens per query than iterative baselines. This result underscores the algorithmic advantage of our approach, which reorganizes the corpus into atomic QA pairs and decouples retrieval from LLM reasoning.

To explore the upper performance bound of our system, we further evaluate a version where the atomic QA knowledge base is constructed offline using the more powerful \textsc{GPT-4} model. This configuration leads to improved accuracy, as shown in Table~\ref{tab:main_results}. It is important to clarify that this preprocessing step incurs a one-time, The online inference stage still remains fully based on \textsc{LLaMA3.1-8B}.

In summary, these findings demonstrate that \textsc{CompactRAG} delivers a favorable balance of efficiency and accuracy even with a middle scale LLM. Furthermore, the system's accuracy exhibits potential for further enhancement through improvements in the quality of the underlying atomic QA knowledge base.
\subsection{Token Efficiency Analysis}

We further evaluate the token efficiency of \textsc{CompactRAG} in comparison with several iterative retrieval–reasoning baselines.  
Figure~\ref{fig:cumulative_token} shows the cumulative token consumption on \textsc{HotpotQA} (others are shown in Appendix~\ref{appdeix:token}) as the number of user queries increases.  
Because \textsc{CompactRAG} includes an offline preprocessing to construct the atomic QA knowledge base, it incurs an initial token cost before online inference begins.  
However, as the number of user requests grows, the cumulative cost curve of \textsc{CompactRAG} increases at a much slower rate than those of iterative RAG baselines such as \textsc{Self-Ask}, \textsc{IRCoT}, and \textsc{Iter-RetGen}.  
The initial offline expense is quickly amortized, and the overall token usage remains substantially lower than that of the iterative methods.  
This result demonstrates the long-term efficiency of \textsc{CompactRAG}, particularly in deployment scenarios involving large volumes of user interactions.

\begin{figure}[htbp]
    \centering
    \includegraphics[width=\linewidth]{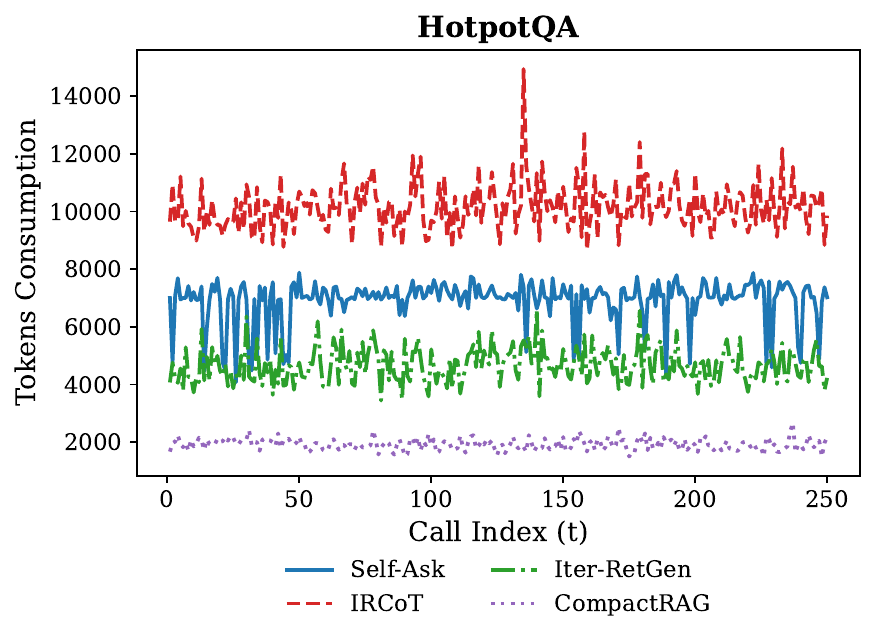}
    \caption{
    \textbf{Per-query token consumption on \textsc{HotpotQA}.}
    Each point represents one user query (a multi-hop question).  
    The token cost varies with question complexity, leading to oscillations across the curve.  
    Despite this variation, \textsc{CompactRAG} maintains consistently lower per-query consumption than iterative baselines, reflecting its efficiency and stability in online inference.
    }
    \label{fig:percall_token}
\end{figure}

To provide a more granular view, Figure~\ref{fig:percall_token} plots the token consumption per user query.  
Here, the horizontal axis corresponds to the sequence of user queries (each representing a distinct multi-hop question), and the vertical axis denotes the total tokens consumed to resolve the query.  
The observed fluctuations are primarily due to differences in question complexity, queries requiring deeper reasoning or more hops naturally consume more tokens.  
Nonetheless, \textsc{CompactRAG} consistently maintains a much lower average token cost across queries compared to iterative baselines.  
This stability results from its fixed two-call design and compact QA retrieval, which together eliminate redundant LLM invocations while preserving reasoning completeness.
\begin{figure}[htbp]
    \centering
    \includegraphics[width=\linewidth]{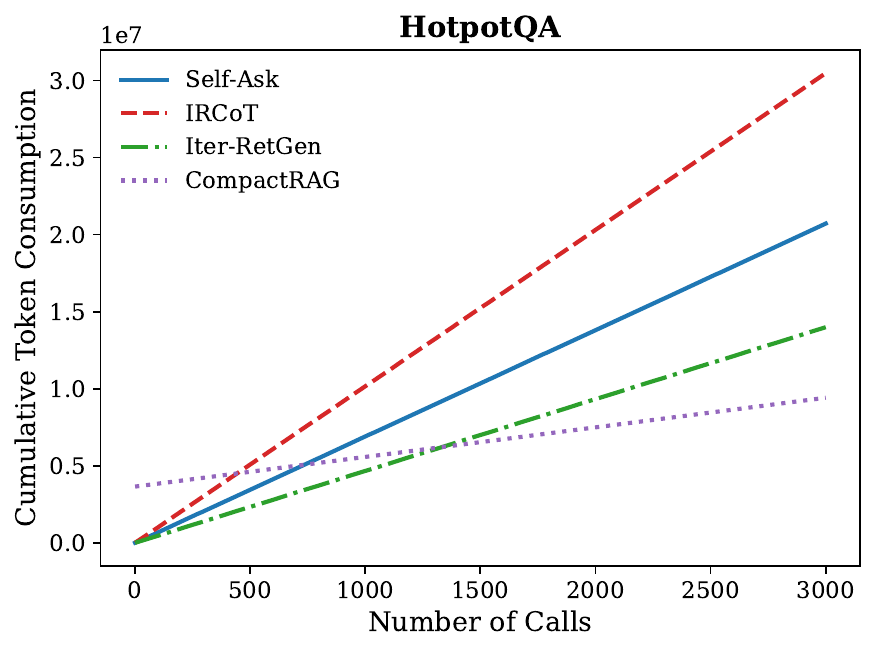}
    \caption{
    \textbf{Cumulative token consumption on \textsc{HotpotQA}.}
    Although \textsc{CompactRAG} incurs an initial offline cost to construct the atomic QA knowledge base, its cumulative token usage grows slowly and eventually remains well below that of iterative baselines as user queries accumulate.
    }
    \label{fig:cumulative_token}
\end{figure}

\subsection{Ablation Study}
\label{sec:ablation}

To evaluate the contribution of the \textit{Answer Extractor} and \textit{Sub-Question Rewriter} modules, we conduct two ablation experiments using the \textsc{LLaMA3.1-8B}-based QA knowledge base. All configurations adopt identical retrieval settings and inference procedures.

\begin{itemize}[leftmargin=1.2em, topsep=2pt]
    \item \textbf{w/o Rewriter:} The rewriter module is removed. The extracted answer is directly concatenated with the next sub-question, encoded by \textsc{Contriever}, and used to retrieve QA pairs.
    \item \textbf{w/o Extractor \& Rewriter:} Both modules are removed. The raw sub-questions generated by the LLM decomposition are encoded by \textsc{Contriever} and used directly for retrieval without any local reasoning.
\end{itemize}

Table~\ref{tab:ablation} reports the accuracy across three benchmarks.  
Removing either component leads to a consistent decline in performance, confirming that both modules are essential for maintaining entity grounding and retrieval precision.  
The degradation is particularly evident when the rewriter is removed, indicating that explicit entity resolution is critical for accurate multi-hop reasoning.

\begin{table}[t]
\centering
\caption{Ablation results (Accuracy \%) on three benchmarks using the \textsc{LLaMA3.1-8B} QA knowledge base.}
\label{tab:ablation}
\renewcommand{\arraystretch}{1.05}
\setlength{\tabcolsep}{4pt} 
\small 
\begin{tabular}{lccc}
\toprule
\textbf{Method} & \textbf{HotpotQA} & \textbf{2WikiMultiHopQA} & \textbf{MuSiQue} \\
\midrule
\textbf{CompactRAG (Full)} & \textbf{70.4} & \textbf{53.2} & \textbf{41.2} \\
w/o Rewriter & 63.2 & 48.8 & 35.8 \\
w/o Extractor \& Rewriter & 58.4 & 44.2 & 32.6 \\
\bottomrule
\end{tabular}
\vspace{-2mm}
\end{table}

These results demonstrate that both the extractor and rewriter modules significantly enhance \textsc{CompactRAG}'s ability to preserve contextual consistency and reasoning accuracy while keeping inference efficient with minimal LLM calls.

\subsection{Discussion}
\label{sec:discussion}

The experimental findings collectively highlight the efficiency and accuracy trade-off addressed by \textsc{CompactRAG}.  
Unlike prior iterative RAG systems, which repeatedly alternate between retrieval and LLM reasoning, our design constrains the number of LLM invocations to two per query while maintaining competitive accuracy.  This fixed call structure not only reduces inference cost but also simplifies the overall pipeline, making it more predictable and scalable for real world deployment. The results also reveal that the quality of the atomic QA knowledge base plays a crucial role in downstream reasoning.  
When the QA base is constructed using a stronger reader such as \textsc{GPT-4}, accuracy improves across all benchmarks, demonstrating that enhancing the semantic fidelity of offline knowledge can directly boost reasoning quality at inference time.  
However, even with a smaller LLM such as \textsc{LLaMA3.1-8B}, \textsc{CompactRAG} achieves comparable performance to much heavier iterative methods, underscoring the robustness of its design.

From a broader perspective, these results suggest that efficient reasoning in retrieval-augmented systems does not necessarily require larger models or more frequent model calls.  
Instead, structuring external knowledge into concise, semantically aligned units and leveraging lightweight reasoning components can yield comparable accuracy with drastically lower computational overhead.  
This insight opens a promising direction for future research on scalable, cost-efficient.




\section{Conclusion}
\label{sec:conclusion}

This paper introduced \textsc{CompactRAG}, a retrieval-augmented generation framework designed to achieve efficient multi-hop reasoning with minimal LLM usage.  
By decoupling retrieval and reasoning through the construction of an atomic QA knowledge base and the integration of lightweight reasoning modules, \textsc{CompactRAG} reduces token consumption and stabilizes inference cost regardless of question complexity.  
Unlike iterative RAG methods that repeatedly invoke LLMs, our approach fixes the number of calls to two per query while maintaining competitive accuracy across multiple benchmarks. Extensive experiments on \textsc{HotpotQA}, \textsc{2WikiMultiHopQA}, and \textsc{MuSiQue} demonstrate that \textsc{CompactRAG} significantly lowers computational overhead without sacrificing answer quality.  
Ablation studies further confirm the complementary roles of the \textit{Answer Extractor} and \textit{Sub-Question Rewriter}, while additional analysis show that improving the semantic quality of the atomic QA base can further enhance performance.  

Overall, \textsc{CompactRAG} highlights a promising direction for developing cost-efficient and scalable RAG systems.  
By combining modular reasoning, efficient retrieval, and pre-processed knowledge, it offers a practical blueprint for large-scale multi-hop reasoning tasks.  
Future work will explore adaptive retrieval strategies, dynamic sub-question generation, and cross-domain generalization of QA knowledge bases, extending the framework to broader open-domain reasoning, interactive dialogue, and knowledge-intensive NLP applications.

\begin{acks}
This work was supported by the National Natural Science Foundation of China (Grant No. 62306138), the Jiangsu Natural Science Foundation (Grant No. BK20230784), and the Innovation Program of the State Key Laboratory for Novel Software Technology at Nanjing University (Grant Nos. ZZKT2024B15 and ZZKT2025B25).
\end{acks}

\bibliographystyle{ACM-Reference-Format}
\balance
\bibliography{sample-base}
\appendix
\onecolumn

\section{QAs Generation Prompt}
\label{prompt:qa}
\begin{tcolorbox}[title=Generate QAs Prompt]
\small
\textbf{System Role:} Knowledge extraction and question generation system. \\[4pt]
\textbf{Task Description:} You will receive: 
\begin{enumerate}[nosep,leftmargin=1.2em]
    \item Original passage text
    \item Extracted entities and relationships from the passage
\end{enumerate}

Your task is to generate atomic knowledge facts and corresponding QA pairs. \\[4pt]

\textbf{Output Format:}  
A single JSON object only (nothing else) enclosed in a \texttt{```json ... ```} code block with exactly two keys:
\begin{itemize}[nosep,leftmargin=1.2em]
    \item \texttt{"atomic\_facts"}: an array of atomic knowledge facts (each fact should be independent, complete, and self-contained)
    \item \texttt{"qa"}: an array of objects \{\texttt{"question": "...", "answer": "..."}\}
\end{itemize}

\textbf{Rules for Atomic Facts:}
\begin{enumerate}[nosep,leftmargin=1.2em]
    \item Each fact should be an independent, complete statement that can stand alone.
    \item Cover both explicit and implicit relationships mentioned in the text.
    \item Include background knowledge and context that help understand the entities.
    \item Each fact should be concise but informative (preferably one sentence).
    \item Do not duplicate facts or add information not present in the passage.
\end{enumerate}

\textbf{Rules for QA Generation:}
\begin{enumerate}[nosep,leftmargin=1.2em]
    \item Each question must be short ($\le$ 12 words) and start with a question word (\textit{Who, What, When, Where, Which, How, How many}).
    \item Use explicit entity names from the entities list; avoid pronouns or vague references.
    \item Each answer must be an exact verbatim substring from the original passage.
    \item Ensure coverage of all important entities and relationships.
    \item Avoid duplicate questions or answers.
\end{enumerate}

\textbf{Example:}

\textit{[Original Text]:} \\
\smallskip
\texttt{Lilli's Marriage (German: Lillis Ehe) is a 1919 German silent film directed by Jaap Speyer. It is a sequel to the film "Lilli", and premiered at the Marmorhaus in Berlin. The film's art direction was by Hans Dreier.}

\textit{[Entity List]:} \\
\smallskip
\texttt{Lilli's Marriage (WORK\_OF\_ART), Lillis Ehe (WORK\_OF\_ART), Jaap Speyer (PERSON), Lilli (WORK\_OF\_ART), Marmorhaus in Berlin (FAC), Hans Dreier (PERSON), 1919 (DATE)}

\textit{[Output JSON]:}

\begin{verbatim}
```json
{
  "atomic_facts": [
    "Lilli's Marriage is a 1919 German silent film",
    "Lilli's Marriage is also known as Lillis Ehe in German",
    "Jaap Speyer directed Lilli's Marriage",
    "Lilli's Marriage is a sequel to the film Lilli",
    "Lilli's Marriage premiered at the Marmorhaus in Berlin",
    "Hans Dreier was responsible for the art direction of Lilli's Marriage",
    "Lilli's Marriage was released in 1919"
  ],
  "qa": [
    {"question": "What is Lilli's Marriage?", "answer": "a 1919 German silent film"},
    {"question": "Who directed Lilli's Marriage?", "answer": "directed by Jaap Speyer"},
    {"question": "Which film is Lilli's Marriage a sequel to?", "answer": "It is a sequel to the film \"Lilli\""},
    {"question": "Where did Lilli's Marriage premiere?", "answer": "premiered at the Marmorhaus in Berlin"},
    {"question": "Who was responsible for the art direction of Lilli's Marriage?", "answer": " Hans Dreier"}
  ]
}
\end{verbatim}

\textit{Now process the following passage:}
\begin{verbatim}
[Original Text]:
{passage}

[Entity List]:
{entity_info}

\end{verbatim}
\end{tcolorbox}



\section{Prompt-evaluate answer based LLM}
\label{prompteval}
\begin{tcolorbox}[title=Prompt-evaluate answer based LLM]
You are an experienced linguist who is responsible for evaluating the correctness of the generated responses.\\
You are provided with question, the generated responses and the corresponding ground truth answer.\\
Your task is to compare the generated responses with the ground truth responses and evaluate the correctness of the generated responses.
Response directly ``yes'' or ``no''.\\

Question: \{question\}\\
Prediction: \{prediction\}\\
Ground-truth Answer: \{answer\}\\
Your response:
\end{tcolorbox}

\section{Additional Token Efficiency Results on Other Benchmarks}
\label{appdeix:token}

To further validate the token efficiency of \textsc{CompactRAG}, we report additional analyses on the \textsc{2WikiMultiHopQA} and \textsc{MuSiQue} benchmarks.  
Each benchmark includes two visualizations: cumulative token consumption and per-query token consumption.  
Both exhibit the same trends as observed on \textsc{HotpotQA}—an initial offline cost for building the QA knowledge base, followed by sustained efficiency during online inference.  
These consistent patterns demonstrate that \textsc{CompactRAG} maintains its efficiency advantage across different datasets and reasoning complexities.
\begin{figure*}[htbp]
    \centering
    \begin{subfigure}[t]{0.35\textwidth}
        \centering
        \includegraphics[width=\linewidth]{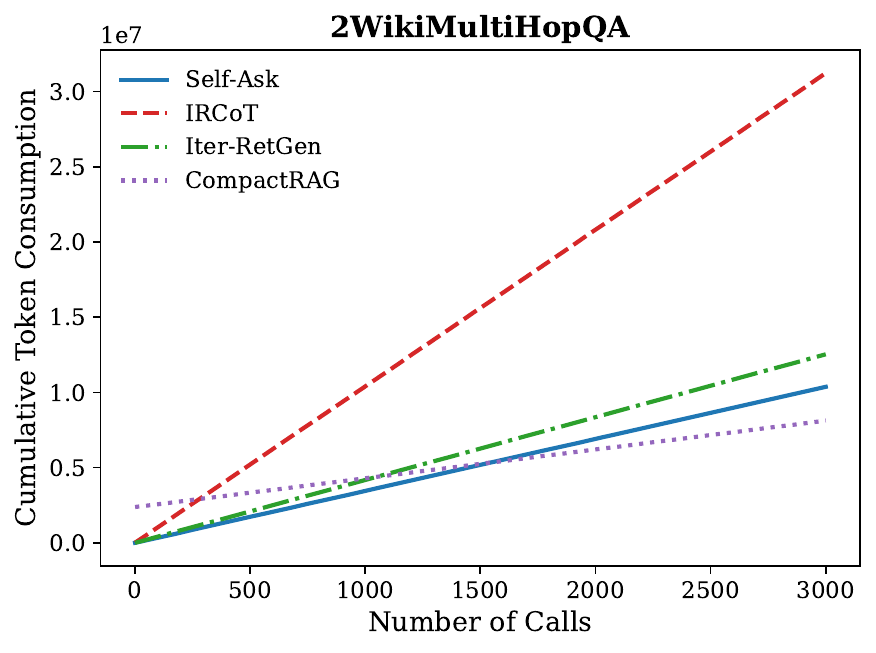}
        \label{fig:2wiki_cumulative}
    \end{subfigure}
    \hfill
    \begin{subfigure}[t]{0.35\textwidth}
        \centering
        \includegraphics[width=\linewidth]{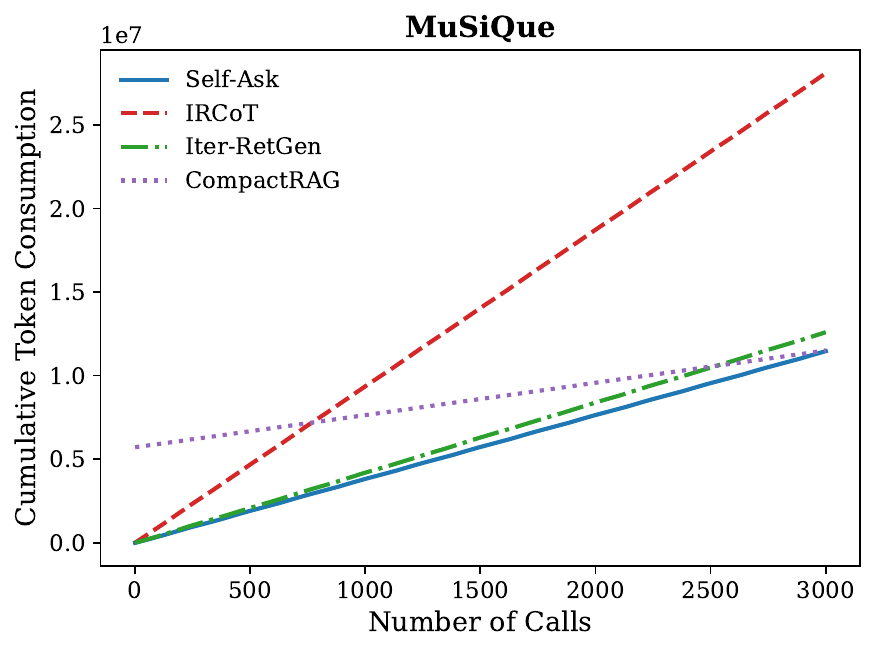}
       
        \label{fig:musique_cumulative}
    \end{subfigure}
    \hfill
     \begin{subfigure}[t]{0.35\textwidth}
        \centering
        \includegraphics[width=\linewidth]{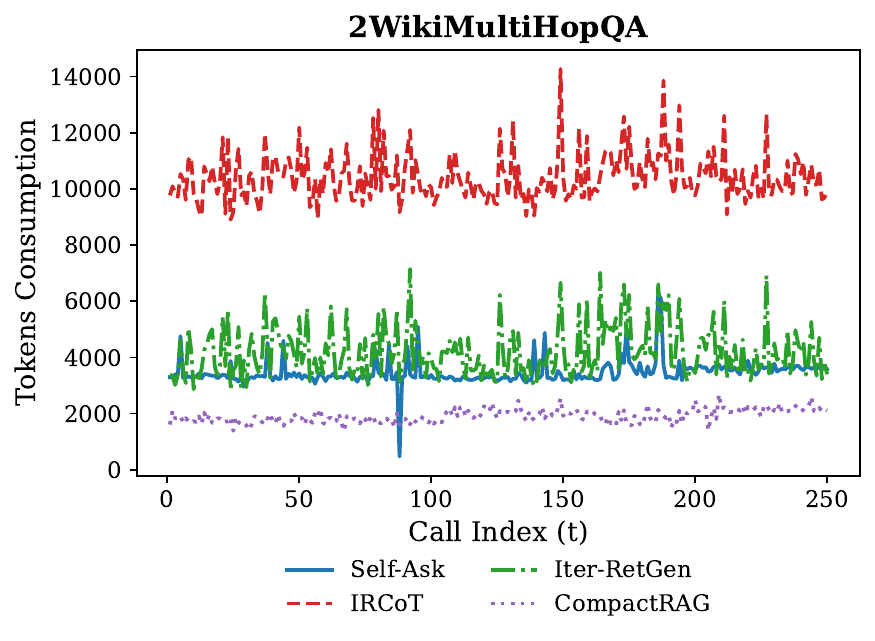}
        \label{fig:2wiki_percall}
    \end{subfigure}
    \begin{subfigure}[t]{0.35\textwidth}
        \centering
        \includegraphics[width=\linewidth]{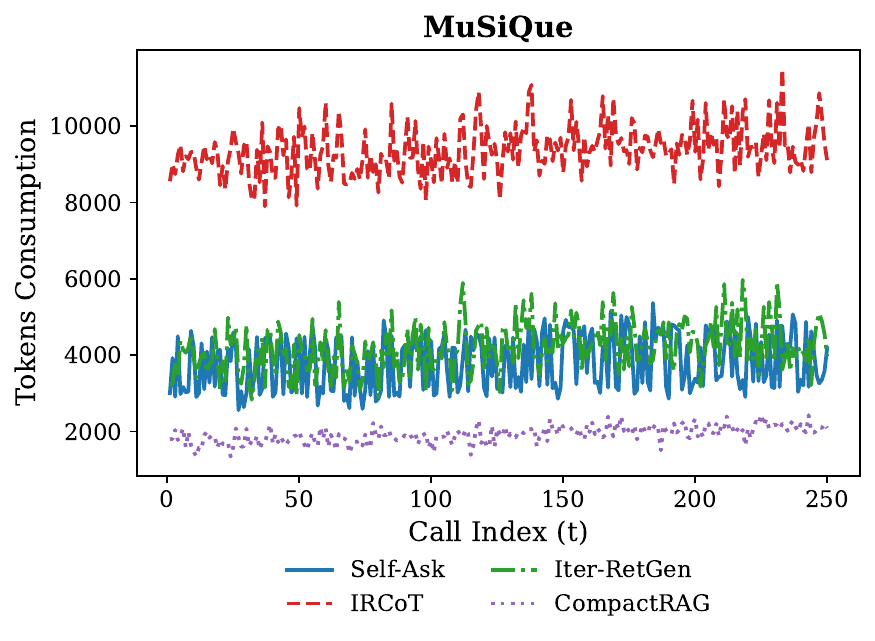}
       
        \label{fig:musique_percall}
    \end{subfigure}
    \caption{Token consumption comparison across additional benchmarks.
    }
    \label{fig:appendix_token_analysis}
\end{figure*}

\end{document}